\documentclass{article}
\usepackage{arxiv}
\usepackage{multirow}
\usepackage[utf8]{inputenc} 
\usepackage[T1]{fontenc}    
\usepackage{hyperref}       
\usepackage{url}            
\usepackage{booktabs}       
\usepackage{amsfonts}       
\usepackage{nicefrac}       
\usepackage{microtype}      
\usepackage{lipsum}		    
\usepackage{graphicx}
\usepackage{amsmath}
\usepackage{arydshln}
\usepackage{booktabs}
\usepackage{natbib}
\usepackage{doi}
\usepackage{stackengine}
\usepackage{CJKutf8}
\usepackage{array}
\usepackage{xcolor}
\usepackage{marvosym}
\usepackage{natbib}
\setcitestyle{numbers,square}
\newcommand\xrowht[2][0]{\addstackgap[.5\dimexpr#2\relax]{\vphantom{#1}}}

\title{{\bf{ChineseWebText}}: Large-Scale High-quality Chinese Web Text Extracted with Effective Evaluation Model}

\date{} 					

\author{
Jianghao Chen$^{1,2}$\thanks{Equal Contribution.} \quad
Pu Jian$^{1,2}$~\footnotemark[1] \quad
Tengxiao Xi$^{1,2}$~\footnotemark[1] \quad
Dongyi Yi$^{3}$~\footnotemark[1] \quad
Qianlong Du $^{1}$~\footnotemark[1] \quad \\
\bf{Chenglin Ding} $^{3}$ \quad
\bf{Guibo Zhu} $^{1,2,3}$\textsuperscript{\Letter}~ \quad
\bf{Chengqing Zong} $^{1,2}$ \quad
\bf{Jinqiao Wang} $^{1,2,3}$\quad
\bf{Jiajun Zhang} $^{1,2,3}$\textsuperscript{\Letter}~ \quad
\\
\small{$^1$ Institute of Automation, Chinese Academy of Sciences }\quad \\
\small{$^2$ School of Artificial Intelligence, University of Chinese Academy of Sciences }\\
\small{$^3$ Wuhan AI Research }\\
\texttt{\{chenjianghao2022,jianpu2023,xitengxiao2022\}@ia.ac.cn}\\
\texttt{\{qianlong.du,gbzhu,cqzong,jqwang,jjzhang\}@nlpr.ia.ac.cn}\\
}

\begin{document}
\maketitle

\begin{abstract}

    During the development of large language models (LLMs), the scale and quality of the pre-training data play a crucial role in shaping LLMs' capabilities. To accelerate the research of LLMs, several large-scale datasets, such as C4 \cite{2020T5C4}, Pile \cite{2020_pile}, RefinedWeb \cite{2023refinedweb} and WanJuan \cite{2023wanjuan}, have been released to the public. However, most of the released corpus focus mainly on English, and there is still lack of complete tool-chain for extracting clean texts from web data. Furthermore, fine-grained information of the corpus, e.g. the quality of each text, is missing. To address these challenges, we propose in this paper a new complete tool-chain \textbf{EvalWeb} to extract Chinese clean texts from noisy web data. First, similar to previous work, manually crafted rules are employed to discard explicit noisy texts from the raw crawled web contents. Second, a well-designed evaluation model is leveraged to assess the remaining relatively clean data, and each text is assigned a specific quality score. Finally, we can easily utilize an appropriate threshold to select the high-quality pre-training data for Chinese. Using our proposed approach, we release the largest and latest large-scale high-quality Chinese web text \textbf{ChineseWebText}, which consists of 1.42 TB and each text is associated with a quality score, facilitating the LLM researchers to choose the data according to the desired quality thresholds. We also release a much cleaner subset of 600 GB Chinese data with the quality exceeding 90\%. The data, codes and the tool-chain are available in this website \footnote{\url{https://github.com/CASIA-LM/ChineseWebText}}.
    
   
\end{abstract}

\section{Introduction}

Recent years have witnessed the rapid progress of large language models (LLMs). The models, such as GPT-3\cite{brown2020language}, BLOOM\cite{scao2022bloom}, LLaMA\cite{llama}, Falcon\cite{2023refinedweb}, PaLM\cite{chowdhery2022palm} and GPT-4\cite{GPT4}, become more and more powerful, even performing better than humans in some natural language understanding and generation tasks.  
During the development, it is evident that the scale and quality of pre-training data play a crucial role on LLM's capability. A large-scale and high-quality dataset is the foundation of LLMs and is the source of all the LLM's amazing capabilities.


 
In order to expedite the research on LLMs, several large-scale datasets have been made publicly available in recent years, such as C4 \cite{2020T5C4}, Pile \cite{2020_pile}, RefinedWeb \cite{2023refinedweb} and WanJuan \cite{2023wanjuan}. Previous studies usually collect the raw texts at first from various sources, such as Wikipedia, GitHub, ArXiv, Stack Exchange, and CommonCrawl, in which CommonCrawl data often accounts for the vast majority. Then, handcrafted rules are designed to filter out the raw data in three steps: extracting the data in the language of interested, filtering out the noisy texts with language-specific rules and data deduplication. It should be noted that, most of the previous studies mainly focus on the collection of English-centered texts, and there is lack of a complete tool-chain for extracting clean data centered in other languages, e.g. Chinese. Furthermore, previous work usually directly release the final data, without giving the fine-grained information of the text, such as the quality of each text, limiting the potential that assists LLM researchers to re-filter the data according to their desired quality threshold.

To address these problems, in this paper we introduce a new complete tool-chain \textbf{EvalWeb}, which could extract high-quality Chinese texts from raw web data. The whole process can be divided into two parts. The first part is similar to previous studies and mainly utilizes manually designed rules to filter out explicit noisy data, generating the initial Chinese clean data. This part processes the web texts with two modules: preparation module and processing module. The preparation module first employs a language identification model to extract Chinese data, and then adopts a hash-based deduplication algorithm to remove duplicate texts. The preprocessing module then handles the resulting data with well-designed rules, including length filtering, sensitive words filtering and filtering with Chinese character ratio. Previous studies usually stop after the usage of these rules. In contrast, we introduce the second part which is a quality evaluation module. Due to the diversity of web texts, the remaining dataset after the usage of filtering rules still contains a large number of low-quality text, which cannot be cleared using manually crafted rules. Consequently, we propose to design a BERT-based quality evaluation model for assessing all the remaining data, thereby generating a quality score for each text. Finally, we can select the high-quality Chinese data in the dataset according to the quality threshold. Using our complete tool-chain \textbf{EvalWeb}, we release the latest and largest Chinese dataset \textbf{ChineseWebText}, which consists of 1.42 TB data and each text is assigned a quality score, facilitating LLM researchers to select data according to a new quality threshold. We also release a much cleaner subset of 600 GB Chinese texts with quality exceeding 90\%.

Our contributions can be summarized as follows:

(1) In this paper, we propose a new complete tool-chain \textbf{EvalWeb}, which could extract high-quality Chinese pre-training data from noisy web texts.

(2) In this paper, we release the latest and largest Chinese dataset consisting of 1.42 TB, and each text in this dataset is assigned a quality score according to our quality evaluation module. We further release a much cleaner subset of 600 GB Chinese texts with quality exceeding 90\%.


\section{Related Work}

\textbf{Rule-based Text Filtering.} Rule-based text filtering methods are the dominant paradigm to identify content-rich and semantically coherent data from collected raw datasets with handcrafted rules. During the collection of pre-training data, there are a large number of text data on the web. However, these data include a lot of noise, such as violence, pornographic, advertisement and error characters. Consequently, in order to extract high-quality data, several rule-based methods have been proposed to explore how to automatically filter undesired content from noisy web data. In these work, deduplication\cite{lee_deduplicating_2022} methods are employed to remove duplicate text from the data, while some handcrafted rules \cite{raffel_exploring_2020,luccioni_whats_2021} are adopted to filter out violence, pornographic, advertisement and other explicit noisy data. Besides, perplexity \cite{wenzek_ccnet_2020} is also usually used to evaluate the fluency of the texts. However, these work mainly focus on English and lack a complete tool-chain for Chinese.

\textbf{Text Classification Model}. Different from rule-based text filtering methods, text classification model is an alternative approach to identify high-quality data with a well-designed classifier. The simplest text classification model is logistic regression\cite{brown2020language}, which uses the logistic function to calculate the probability values for each text, and then classifies them into positive or negative with a designed threshold. Currently, BERT\cite{devlin_bert_2018} and FastText\cite{joulin_bag_2017} are both commonly used text classification models. BERT is a transformer-based\cite{transformer} pre-training language model that has achieved remarkable performance in various text classification and understanding tasks. Through pre-training on masked language model and next sentence prediction tasks with a large dataset, this model learns powerful language understanding and representation abilities, which makes it perform well on text classification tasks. FastText\cite{joulin_bag_2017} is also a neural network based approach which is similar to CBOW\cite{word2vec}. It is characterized by its ability to train efficiently and quickly on large-scale data, while achieving competitive classification performance. In this paper, both of these two approaches will be employed to evaluate the qualities of the Chinese texts.  

\textbf{Datasets for Pre-training.} In recent years as the scale of pre-trained language models expands, there is a concomitant increase in the demand for large-scale pre-training datasets. Due to the convenience of acquisition and cost-efficiency associated with web-scraped data, it has progressively emerged as a pivotal source for pre-training datasets\cite{raffel2020exploring}. In these work, Gao et al. (2020) \cite{gao2020pile} build a 825 GB English corpus by mixing established natural language processing datasets and several newly introduced ones. This dataset covers 22 diverse high-quality subsets which derive from academic or professional sources, including PubMed Central, the FreeLaw Project, Stack Exchange, Books3\cite{presser2020books3}, OpenSubtitles\cite{tiedemann2016finding} and so on. Different from the work of Gao et al. (2020) \cite{gao2020pile}, Penedo et al. (2023)\cite{penedo2023refinedweb} demonstrate that properly filtered and deduplicated web data alone can also train a powerful model, even outperforming the LLMs trained on curated corpora. They use a pipeline approach to filter and deduplicate web data from CommonCrawl at very large scale and then release an English dataset which have 600 billion tokens. In addition, with data from web and synthetically generated textbooks and exercises with GPT-3.5, Gunasekar et al. (2023)\cite{gunasekar2023textbooks} build a code dataset which has 7B tokens. They mainly focus on the coding functions of LLMs, esecpically the writing Python functions. Moreover, He et al. (2023) \cite{he2023wanjuan} release a comprehensive multimodal dataset, which also contains texts in both Chinese and English, and is collected from a wide range of web sources. These public datasets mainly focus on English, and lack a complete tool-chain for extracting Chinese clean data from web sources. Furthermore, they only contain the cleaned texts, while missing the corresponding fine-grained information (e.g. the quality of each text) which could help LLM researchers to re-filter texts with new desired quality thresholds.

\section{Data Construction}

\begin{figure}[htbp]
  \centering
  \includegraphics[width=1\textwidth]{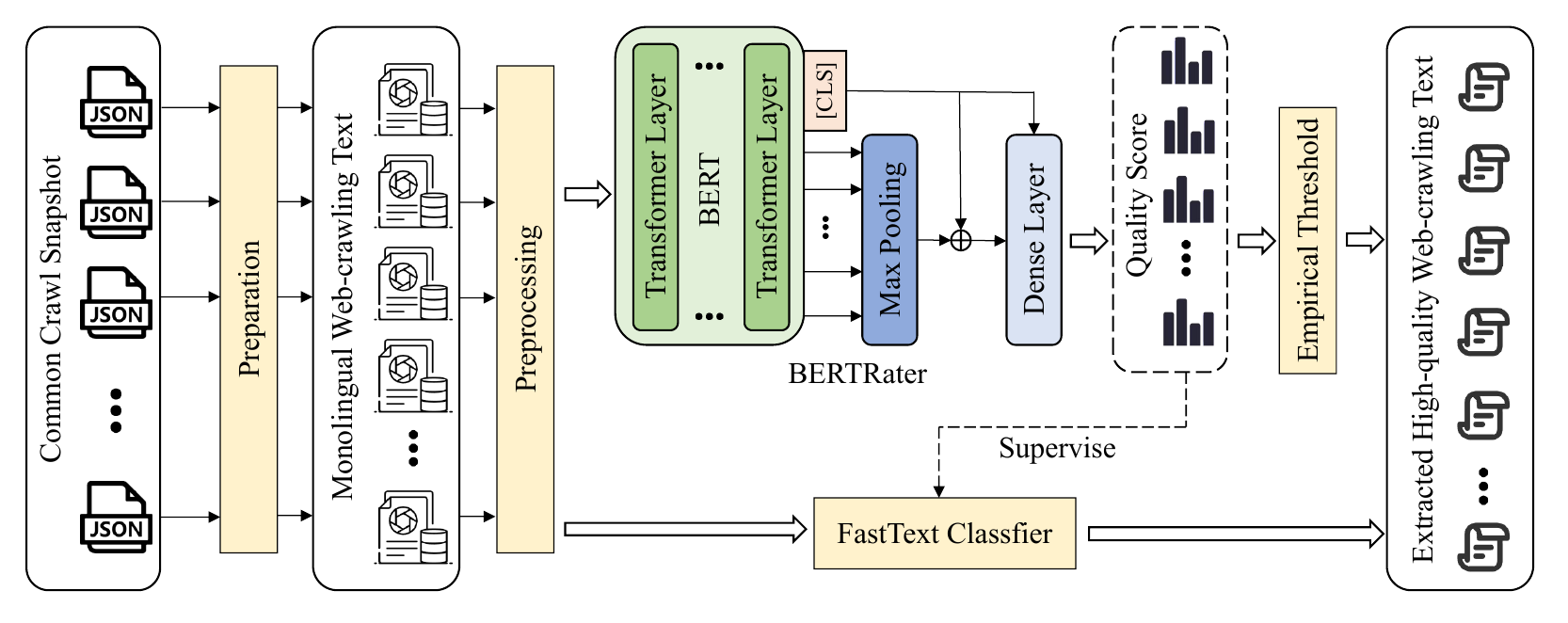}
  \caption{The architecture of our EvalWeb approach.}
  \label{fig1}
\end{figure}

Due to the presence of substantial noise and irrelevant information on the web, extracting high-quality Chinese data from the web poses a significant challenge. In order to extract high-quality Chinese text from web effectively, in this paper we propose a pipeline system \textbf{EvalWeb}, which integrates mannual crafted rules and evaluation models. With this approach, we can effectively filter undesirable content such as offensive speech, advertisements and idle chatter, and then extract high-quality Chinese texts. As in Figure \ref{fig1}, it illustrates the overview of our proposed approach. For the crawled data from web, we first use a preparation module to process them, and then extract the monolingual Chinese data. After that, a preprocessing module will be used to further filter them with mannual crafted rules, including data length, sensitive words, proportion of Chinese characters and so on. Finally, a BERT-based evaluation model will be employed to assess the qualities of filtered data. By this way, we can generate a quality score for each of the text, and then use an appropriate threshold to extract the high-quality data as we required. Furthermore, 
considering computational cost and efficiency, we further propose to leverage knowledge distillation\cite{stanton2021does} techniques to train a FastText classifier, which can achieve similar performance with faster efficiency and lower computational costs.

\subsection{Data Collection and Preparation}

As a publicly accessible web scrape dataset, CommonCrawl has been running for 12 years and has accumulated petabytes of web data. Consequently, we regard it as the source of our web data. In this paper, we collect nine latest CommonCrawl snapshots from the internet, including "2021-43", "2022-05", "2022-21", "2022-27","2022-33","2022-49","2023-06","2023-14" and "2023-23". These obtained snapshots are compressed plain text, each of which is 8-10 TB in size (approximately 3 billion web pages). Each snapshot is regrouped into JSON format shards of 5 GB, where each item corresponds to a web page. 

Since the original CommonCrawl file is typically quite diverse and contains texts in many languages, to efficiently extract Chinese text data from it, we first employ a deduplication and language identification (LID) module to perform preliminary cleaning on the collected datasets. Following the work of CCNet\cite{wenzek_ccnet_2020}, in this module a Hash-based inter-string deduplication method is employed to remove duplicate text from different snapshots. Additionally, a well-trained language identification model\cite{grave2018learning}, which could support 157 languages, is applied to select Chinese data. By this way, we can obtain all the monolingual Chinese text data we required. 

\subsection{Preprocessing}

\begin{table}[htbp]
    \caption{ Examples for different filtering rules.}\label{Data Filtering Statistics}
    \centering
    \begin{tabular}{p{4cm} | p{9cm} }
    \toprule
     Filtering Operation & Example \\
     \midrule
     \multicolumn{1}{m{4cm}|}{Text Extraction}  & \multicolumn{1}{m{9cm}}{\begin{CJK}{UTF8}{gbsn} \{"url": "http://sarahokane.com/ywly/index.aspx",\newline "date\_download": "2022-05-28T10:17:39Z", \newline "length": 854, \newline "nlines": 17, \newline "source\_domain": "sarahokane.com", \newline "title": "合肥市建设投资控股(集团)有限公司", \newline \textcolor{red}{"raw\_content": " 乡村振兴和现代农业板块\textbackslash n乡村振兴与现代农业...注册资本4.39亿元。",} \newline ...\newline "language": "zh", \newline "bucket": "head"\} \end{CJK}}  \\
     \midrule
     \multicolumn{1}{m{4cm}|}{Length less than 200}  & \multicolumn{1}{m{9cm}}{\begin{CJK}{UTF8}{gbsn} "德国黑森州法兰克福AMazon数据中心" \end{CJK}}  \\
     \midrule
     \multicolumn{1}{m{4cm}|}{Average line length less than 10}  & \multicolumn{1}{m{9cm}}{\begin{CJK}{UTF8}{gbsn} "汽车资讯\textbackslash n汽车制造商\textbackslash n学车租车\textbackslash n俱乐部,汽车网,汽车报价,新车,汽车图片\textbackslash n汽车网,汽车报价,新能源,新车,汽车图片\textbackslash n南方网：汽车频道\textbackslash n维修,改装,车模,新车,用车\textbackslash n ......" \end{CJK}}  \\
     \midrule
     \multicolumn{1}{m{4cm}|}{Traditional Chinese characters}  & \multicolumn{1}{m{9cm}}{\begin{CJK}{UTF8}{bsmi} "有看過裝潢中的便利商店嗎？它可能內部還沒整備好，甚至根本是空盪盪的一片。但店門一定會上紅布條，寫著「XXX便利商店在此為您服務」。去過外縣市旅遊嗎？當你開著車要找某家店時，你大概不會開到正門口才看是不是你要找的店。" \end{CJK}}  \\
     \midrule
     \multicolumn{1}{m{4cm}|}{Proportion of Chinese characters fewer than 30\%}  & \multicolumn{1}{m{9cm}}{\begin{CJK}{UTF8}{gbsn} "线上买球平台\textbackslash u0006\textbackslash u0007、精准\textbackslash u0007\textbackslash u0005、可靠的传感技术解决方案及产品\textbackslash n\textbackslash nXSENS MTi 600系列 ...\textbackslash n\textbackslash n7 GNSS/INS\textbackslash n线上买球平台\textbackslash u0006，具有多个GN......</p>\textbackslash nMTi-G-710 GNSS/I...\textbackslash n<p><span style="font-size: 12px;">......" \end{CJK}}  \\
     \midrule
     \multicolumn{1}{m{4cm}|}{Occurrence of sensitive words more than 0.5 per line}  & \multicolumn{1}{m{9cm}}{\begin{CJK}{UTF8}{gbsn} "宝博体育强化创新引领，坚持“科技宝博体育”战略，构建“以企业为主体、市场为导向、产学研相结合”的科技创新体系。\textbackslash n2019年度大事记\textbackslash n友情链接：玩球直播nba 真钱滚球真人 金花三张牌赢钱 55直播nba 买球 手机轮盘app 亚博app在线登录 亚博app登录......" \end{CJK}}  \\
     \midrule
     \multicolumn{1}{m{4cm}|}{Internal duplication ratio greater than 50\%}  & \multicolumn{1}{m{9cm}}{\begin{CJK}{UTF8}{gbsn} "\textcolor{red}{“山东省民间融资机构宣传月活动”......活动期间吸引了当地市民\textbackslash{}n}2018年11月9日，\textcolor{red}{“山东省民间融资机构宣传月活动”......活动期间吸引了当地市民}的广泛关注，现场发放宣传资料600余份，解答市民疑问100余人次，德州广播电视台，德州日报，德州晚报全程采访报道。" \end{CJK}} \\
     \bottomrule
    \end{tabular}
\end{table}

After getting the monolingual Chinese web data, in this section we will focus on how to extract high-quality Chinese texts from them. Given the prevalence of violent, pornographic, advertising, and error characters in web data, we will first employ some manually crafted rules to filter out these noisy data. The details of these crafted rules are presented in the following.

 \textbf{Text Extraction} After the data preparation stage, there exists a substantial amount of redundant content, which holds little substantive value for subsequent analysis, such as irrelevant key-value pairs. To ensure the accuracy and efficiency of data analysis, we initially undertake the task of extracting all textual content from the entire dataset.


\textbf{Data Length} In web texts, a substantial portion of data consists of documents with short text lines which are separated by '\textbackslash n'. And the texts in different lines do not have significant semantic relevance to each other, which results in that these data are not useful for the training of language models. To eliminate excessively documents with short text lines, we will calculate the average text line length for each document and then remove documents with an average line length of fewer than 10 characters. Besides, during pre-training procedure, short text data usually contains limited information, making it ineffective in providing context and contextual information. Consequently, we will remove texts whose length is less than 200.




\textbf{Proportion of Characters} The objective of this paper is to create a high-quality simplified Chinese dataset sourced from web data. Therefore, we firstly eliminate text data composed of traditional Chinese characters. Additionally, we found that some data exhibit a notably low percentage of Chinese characters. And the rest of these data are filled with some other language characters, non-essential characters, special symbols, irrelevant markers, and so on. These data are unhelpful for the training of large language models. Consequently, we will remove the texts with fewer than 30\% Chinese characters.

\textbf{Sensitive Words} In web data, there is a large amount of harmful texts, including sex, gambling, violence, discrimination, drugs, religion, and so on.  These texts can make large language models generate toxic contents, which have a big negative influence on society, nations and individuals. To avoid these issues, it is necessary to filter out harmful content from the web texts. Firstly, we collect a lot of harmful words and build a sensitive word list. After that, we count the sensitive words in each line of the texts. For one text, if the occurrence of sensitive words exceeds 0.5 per lines, we will regard it as a toxic text and will remove it from our dataset.



\textbf{Internal duplication}
In the training of large language models, duplicate texts can significantly impact training efficiency and model performance. Although we have conducted deduplication in the first stage, subsequent analysis revealed that some duplicate information still exists in the texts. Therefore, we adopt a granularity of 13-gram for analysis, quantifying the proportion of repetitive 13-gram character sequences across all data. When the proportion of repeated 13-gram characters in a data sample exceeds 50\%, we opt to filter it out.



Through the aforementioned rigorous data preprocessing steps, a substantial amount of low-quality data is filtered out. After that, a quality evaluation model will be employed to evaluate the quality scores of the remaining data and then extract the high-quality data with a desired threshold. 

\subsection{Quality Evaluation}
\label{quality_evaluation}
\subsubsection{BERTEval}

In preprocessing procedure, we have used some handcrafted rules to remove the explicit noisy texts from our dataset. However, within the remaining data, there is still a considerable amount of low-quality text data, which cannot be filtered out with handcrafted rules. In order to extract the data of higher quality from them, in this section we further propose to design an evaluation model. In this approach, we will develop a BERT-based classification model to generate a quality score for each text, and then filter the high-quality data with a threshold. The details of the classification model are presented in the following.


\textbf{Training Data Composition} While the evaluation in our current experiment targets CommonCrawl data, we believe the positive training samples should encompass a variety of text types, such as Wikipedia, e-books, poetry, news, and Q\&A data, to prevent the model from exhibiting bias toward deeming any specific text type as high quality. Since CommonCrawl data has a relatively high noise level overall, we directly sampled from CommonCrawl and used the sampling results as negative examples. Table \ref{bert_data} presents the detailed composition and quantity of the training data.
\begin{table}[htbp]
	\caption{Composition of BERTEval Training Data.}\label{bert_data}
	\centering
	\begin{tabular}{llc}
		\toprule
		Type     & Source     & Size ($\times 10^4 $) \\
		\midrule
		\multirow{8}{*}{Positive samples}   & Wikipedia & 12.50 \\ 
                                            & Sina News & 12.50 \\
                                    		& Cbooks & 12.50  \\
	                                        & Zhihu & 12.40 \\
	                                        & WikiQA & 0.90 \\
	                                        & Law & 0.40 \\
	                                        & Poetry & 0.20 \\
	                                        & GovReport & 0.13 \\
	    \midrule
            \xrowht[()]{10pt}
		Negative sample & CC-sampling & 55.00 \\
		\bottomrule
	\end{tabular}
\end{table}

\textbf{BERTEval Architecture} We utilized Tran-BERT-MS-ML-R\cite{wang_use_2022}, an effective AES model based on the BERT-base architecture, to evaluate the quality of the text obtained from web crawling. To reduce computational complexity, we opted to exclude the sub-document scale representation in Tran-BERT-MS-ML-R, which employs text segmentation at various scales as input. Instead, we focused solely on the text-scale representation, utilizing the $[CLS]$ embedding to extract pertinent information and structural features from the broadest perspective of the text. Simultaneously, the token-scale representation was derived from the sequence outputs of BERT. We anticipate that this token-scale representation will be instrumental in identifying and filtering out texts containing offensive language, sexually explicit terms, and frequent nonsensical vocabulary, typically absent in high-quality corpora. Let $x$ represent the text input. After the application of Max Pooling, the token-scale representation is concatenated with the text-scale representation and passed through a Dense Layer with Sigmoid activation, producing a text quality score $f(x|W)$ that falls within the $(0, 1)$ range.


\textbf{Loss Function} In addition to the MSE loss\cite{mesgar_neural_2018}, we used the following two loss functions: Margin Ranking ($MR$) loss \cite{liu2021temp} and Cosine Similarity ($CS$) loss \cite{wang_use_2022}. Let $D$ denote the CommonCrawl corpus. Each negative sample $x_n$ is sampled from CommonCrawl associated with a fixed low score label $y_n$, which formed $D_n$. The positive sample, $x_p$, denotes curated corpora with ideal quality with a constant score $y_p$, which formed $D_p$. Throughout the training process, the labels for positive and negative samples are persistently $y_p$ and $y_n$, respectively. Moreover, it's noteworthy that a significant portion of high-quality texts is present within the CommonCrawl corpus. Given that the supervision employed is coarse-grained even somewhat inaccurate, it would be imprudent to solely rely on MSE loss to rigidly compel the model to fit these labels. For the $ML$ loss, losses only emerge when the ranking of quality scores for samples within each batch doesn't align with their respective labels. The $CS$ loss evaluates the correlation between quality scores and their supervision, rather than their absolute differences. Therefore, the combined loss is 
\begin{equation}
    \mathcal{L}(\boldsymbol{Y}, f(\boldsymbol{X }| W)) =\alpha MSE(\boldsymbol{Y}, f(\boldsymbol{X} | W))+\beta MR(\boldsymbol{Y}, f(\boldsymbol{X} | W))+ \gamma CS(\boldsymbol{Y}, f(\boldsymbol{X} | W)),
\end{equation}

where $\boldsymbol{Y}$ and $f(\boldsymbol{X} | W)$ are the quality score labels and predictions of a batch, respectively. Given the supervision is coarse-grained, we contend that the combined loss function offers valuable insights for enhancing our capability of BERTEval to assess the relative quality of texts. The BERTEval training process consists of the following two stages, which are shown in Fig 2.

\textbf{Pre-training Stage} At this stage, we extracted positive and negative samples at a 1:1 ratio from the CommonCrawl corpus and the curated corpora. We trained BERTEval based on the loss functions previously described. After this training stage, BERTEval acquired a preliminary ability to discern the quality of web-scraped texts, which will be elaborated in detail in the subsequent experiments section. 

\textbf{Self-training Stage} As mentioned before, there is a considerable proportion of texts with desired quality in the Common corpus $D_n$, which might introduce inaccurate supervision, resulting in neither increasing the epoch nor scaling up the training set leading to a detectable improvement in BERTEval. To ameliorate this problem, we adopt a self-training approach \cite{scudder1965probability}. Let $S_n$ denote a randomly sampled subset of $D_n$. In each self-training iteration, the parameters of BERTEval from the previous iteration, $W^t$, are used to generate the pseudo labels of $S_n$, and then BERTEval is retrained on sampled data in $S_n$ with pseudo labels to learn the new parameters $W^{t + 1}$\cite{mukherjee_uncertainty-aware_2020}. It's worth noting that since the positive samples originate from large-scale, high-reliability, curated corpora that do not require pseudo-labeling, we exclusively sample instances with pseudo-labels being $y_n$. The self-training stage can be formulated as:
\begin{equation}
W^{t+1} = \mathop{\arg\min} \limits_{W} \mathbb{E}_{\boldsymbol{X}_p^l \subset D_p} \mathbb{E}_{S_n \subset D_n} \mathbb{E}_{\boldsymbol{X}_n^l \sim p(x_n| W^{t}), \boldsymbol{X}_n^l \subset S_n}\{\mathcal{L}(\boldsymbol{Y}_p^l \oplus \boldsymbol{Y}_n^l, f(\boldsymbol{X}_p^l \oplus \boldsymbol{X}_n^l | W))\}.
\end{equation}
where $\boldsymbol{X}_p^l$ and $\boldsymbol{X}_n^l$ denote the vectors consisting of $l$ randomly sampled samples from $D_n$ and $S_n$, respectively, and $\boldsymbol{Y}_p^l$ and $\boldsymbol{Y}_n^l$ are the corresponding constant vector labels. Since the output layer is activated by Sigmoid, the quality score $f(x | W)$ can also be regarded as a probability of positive samples $p(y_p | x; W)$. Based on the idea of preferring pseudo-labels with high confidence, in iteration $t$, the probability of selecting each sample $x_n \in S_n$ is denoted by 
\begin{equation}
p(x_n|W^t) = \frac{p(y_n|x_n; W)}{\sum_{x \in S_n} p(y_n|x; W)},
\end{equation}
which is the normalization of $p(y_n|x_n; W)$. Heuristically, we avoid sampling the web-crawled texts where $p(y_n|x_n; W)$ falls within the last $K$ proportions. The value of $K$ is informed by our sampling observations from the CommonCrawl dataset. 


\subsubsection{FastText-based Evaluation Model}

\begin{table}[htbp]
	\caption{Composition of FastText Training Data.}\label{fasttext_data}
	\centering
	\begin{tabular}{llc}
		\toprule
		Type     & Source     & Size ($\times 10^4 $) \\
		\midrule
		\multirow{8}{*}{Positive samples}   & Baike & 20 \\ 
                                            & Cbook & 20 \\
                                    		& Zhidao& 20  \\
		                                    & China News & 20 \\
	                                        & Zhihu & 20 \\
	                                        & WikiQA & 10 \\
	                                        & other news & 10 \\
	                                        & BERT-positive & 40 \\
	    \midrule
            \xrowht[()]{10pt}
		Negative sample & BERT-negative & 160 \\
		\bottomrule
	\end{tabular}
\end{table}

To further enhance data processing efficiency and reduce hardware resource requirements, in this paper we also develop a text evaluation model based on FastText\footnote{https://github.com/facebookresearch/fastText} in addition to BERTEval. FastText is libiary for efficient learning of word representations and text classificaiton. Compared to other classification models, such as SVM, logistic regression, and BERT, FastText could significantly reduce training and inference time while maintaining classification performance. 

In the last section, we have built a BERT-based evaluation model BERTEval which performs a good performance on the quality evaluation of Chinese texts. With this model, we can classify the preprocessed web data into high-quality texts (positive) and low-quality texts (negative). Inspired by the idea of knowledge distillation, we will use these classified texts to guide the training of our FastText model. In our approach, we select 400,000 high-quality texts classified by BERTEval as our positive data, while choosing 1,600,000 low-quality texts as our negative data. In order to increase the diversity of training data, our positive data also include some high-quality Chinese data from some other websites and books, such as Baidubaike, Zhihu, Cbook, ChinaNews and so on. These data have been manually proofread and processed. In this way, we can build a good training dataset with 3200K samples. As shown in Table \ref{fasttext_data}, it presents the composition of our training data.

After collecting these training data, we will use a word segmentation tool to process all the texts, and then input the processed data into FastText to train the model. Through this approach, we can obtain a more efficient quality evaluation model.

 \subsubsection{Evaluation Model Comparison}
 
 In order to compare the performance of different evaluation models, in this section we evaluate them on a test set which includes 300 samples. Here we first list other two baseline quality evaluation models: regression-based approach and perplexity-based approach.
 
 \textbf{Regression-based Evaluator}
 
 Following the work of Gururangan et al. (2022) \cite{gururangan_whose_2022}, we combine logistic regression with a word frequency-based vertorizaiton method to conduct text classification on the testset. In this approach, logistic regression is used to calculate a probability value for each sample, and then a threshold is adopted to determine whether the data point should be classified into positive or negative.

 
 \textbf{Perplexity-based Evaluator}
 
 Perplexity could effectively measure the difficulty of a language model in predicting tokens and reflect the fluency of the input texts. Following the work of Wenzek et al. (2020) \cite{wenzek_ccnet_2020}, we utilize a well-trained language model to calculate the perplexity of the texts and classify them with a threshold based on perplexity values. The samples with lower perplexity values will be classified into positive.
 
 \textbf{Comparison Results}
 
 During testing procedure, we will classify the samples of testset with Regression, Perplexity, BERTEval and FastText models repectively, and then compute the precisions of them on positive data. As in Table \ref{results}, it shows the precisions of different evaluaiton models on the testset. TP represents the number of "True Positive" samples, while FP represents the number of "False Positive" samples. From this table, we can see that our BERTEval evaluaiton model gets a much better performance than the regression and perplexity approaches. Besides, benefiting from the good classified results of our BERTEval model, the FastText-based model could further improve the classification precision. This result indicates that using BERTEval to guide the construction of the FastText-based evaluation model is effective. And with this FastText-based evaluation model, our EvalWeb tool-chain could achieve a better performance while effectively improving processing efficiency and resource utilization.



 \begin{table}[htbp]
 	\caption{Classification results of different evaluation models.}\label{results}
 	\centering
 	\begin{tabular}{lccc}
 		\toprule
 		Model      & Precision(\%) & TP+FP & TP \\
 		\midrule\xrowht[()]{10pt}
 		Regression & 49.57 & 234 & 116 \\
 		\xrowht[()]{10pt}
 		Perplexity &  63.27 & 245 & 155 \\
 		 \xrowht[()]{10pt}
          BERTEval & 73.79 & 103 & 76\\
 		\xrowht[()]{10pt}
        FastText&  \textbf{81.58} & 76 & 62 \\	  
 		\bottomrule
 	\end{tabular}
 \end{table}

\subsection{Quality Control}\label{quality_control}

 \begin{figure}[htbp]
  \centering
  \includegraphics[width=0.9\textwidth]{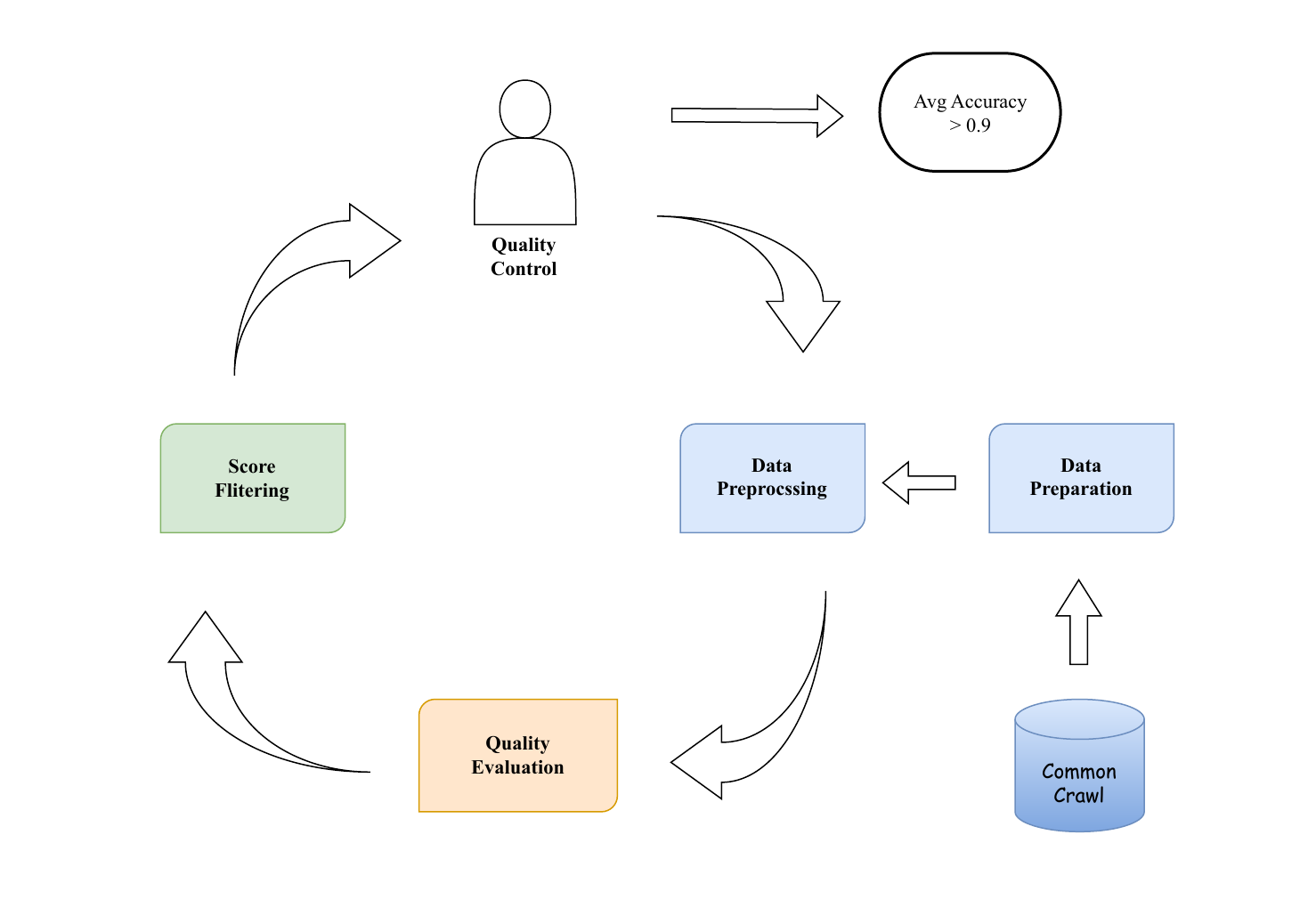}
  \caption{Quality Control.}
  \label{fig2}
 \end{figure}

In section 3.3, after filtering data with a desired quality threshold, we can obtain a Chinese text dataset. In order to ensure the quality of this dataset, We will hire some human evaluators to evaluate its quality. In this method, we will randomly sample 1000 examples from the dataset for three times. After that, three human evaluators will be hired to assess the quality of these data respectively, and the quality of these data is required to be evaluated from the following four aspects: 


\begin{itemize}
\item \textbf{Informativeness}: Whether the text contains enough knowledge and information,  or is just meaningless crap.
\item \textbf{Fluency}: Whether the text has formatting issues, capitalization mistakes, or evident grammatical errors that impair readability.
\item \textbf{Coherence}: Whether the text progressively forms a coherent body of information on a topic through its successive sentences. 
\item \textbf{Toxicity}: Texts used for pre-training should endeavor to exclude offensive remarks, sexually explicit content, and politically sensitive statements to mitigate potential generative risks.
 \end{itemize}

 \begin{table}[htbp]
    \caption{Data samples in json format. The higher score of Sample 1 versus the lower score of Sample 2 demonstrates their differing text qualities, with Sample 1 having better quality than Sample 2. }
    \label{Data Example Format}
    \centering
    \begin{tabular}{p{2.5cm} p{12cm} }
    \toprule
      \textbf{Key} & \textbf{Value}\\
     \midrule
        <\textbf{title}> &\begin{CJK}{UTF8}{gbsn}"潍坊银行2021年上半年净利润同比增长29.57\% 不良率降至1.10\%\_财经\_中国网"\end{CJK}\\
        <\textbf{score}>& 0.95 \\
        <\textbf{text}> & \multicolumn{1}{m{12cm}}{\begin{CJK}{UTF8}{gbsn} "潍坊银行2021年上半年净利润同比增长29.57\% 不良率降至1.10\%\textbackslash n中国网财经8月24日讯 潍坊银行昨日披露2021年二季度信息报告显示，截至2021年6月末，潍坊银行资产总额1920.44亿元，较上年末增长9.34\%；负债总额1789.16亿元，较上年末增长10.54\%。2021年上半年，潍坊银行实现净利润6.09亿元，同比增长29.57\%。\textbackslash n资产质量方面，截至2021年6月末，潍坊银行不良贷款率1.10\%，较上年末下降0.13个百分点。\textbackslash n资本金方面，截至2021年6月末，潍坊银行资本充足率、核心一级资本充足率、一级资本充足率分别为11.66\%、7.89\%、10.13\%，分别较上年末下降1.89、0.89、1.15个百分点。" \end{CJK}}  \\
        <\textbf{url}> & \url{http://finance.china.com.cn/news/special/2021bnb/20210824/5638343.shtml}\\
        <\textbf{source\_domain}> & \url{finance.china.com.cn}\\
         \midrule
        <\textbf{title}> &\begin{CJK}{UTF8}{gbsn}"上海巨也仪器设备有限公司"\end{CJK}\\
        <\textbf{score}>& 0.19 \\
        <\textbf{text}> & \multicolumn{1}{m{12cm}}{\begin{CJK}{UTF8}{gbsn} "石子冲击试验仪\textbackslash n现货提供石子冲击试验机\textbackslash n现货提供石子冲击试验机符合SAE、ASTM、VDA、GM、Ford、Mazda、JIS、Nissan、及Toyota等的测试要求。石子冲击试验机（欢迎实地考察）满足：大众、神龙、通用、日产、马自达、丰田、本田、福特等汽车厂家试验。\textbackslash nNS耐石子冲击性能试验机\textbackslash n耐石子冲击性能试验机主要用于、德系日系、美系汽车厂家试验方法，它能准确再现由飞溅的砂砾造成的破化现象, 适用于外涂层粘聚性破坏试验、涂层系统中不同层间粘合性破坏试验、抗剥落的*涂膜厚度、塑料及玻璃的抗剥落、抗碰撞、抗磨损测试等相关试验。\textbackslash n石子冲击试验机（欢迎实地考察）\textbackslash n石子冲击试验机（欢迎实地考察）符合SAE、ASTM、VDA、GM、Ford、Mazda、JIS、Nissan、及Toyota等的测试要求。满足：大众、神龙、通用、日产、马自达、丰田、本田、福特等汽车厂家试验。\textbackslash n漆膜抗石击试验仪\textbackslash n漆膜抗石击试验仪符合SAE、ASTM、VDA、GM、Ford、Mazda、JIS、Nissan、及Toyota等的测试要求。\textbackslash n石子冲击试验机/抗石子冲击仪\textbackslash n巨也仪器！有大量现货提供，欢迎客户随时来厂参观与指导！" \end{CJK}}  \\
        <\textbf{url}> & \url{http://www.juyesh.com/SonList-1094890.html}\\
        <\textbf{source\_domain}> & \url{www.juyesh.com}\\
        \bottomrule
    \end{tabular}
\end{table}

 During evaluation procedure, each text is assigned a label of either "True" or "False." "True" indicates that the data meets the quality requirement of pre-training in all four aspects, while "False" signifies that the text is noisy to some extent. After completing all the evaluations, we will calculate the average accuracy of these three evaluators. If the average accuracy could exceed 0.9, the filtered Chinese dataset is considered to be a high-quality dataset. Otherwise, we believe that there is still some noisy text in the dataset, and we need to optimize the preprocessing and evaluation modules again, and reprocess the dataset until it meets the quality requirements. The architecture of quality control process is illustrated in Figure \ref{fig2}.

\begin{table}[htbp]
    \caption{Overview of output datasets.}
    \label{total_remain_size}
    \centering
    \begin{tabular}{cp{2cm}p{2cm}p{2cm}}
    \toprule
    \multirow{3.5}{*}{Snapshot}   & \multicolumn{3}{c}{Data Size(GB)} \\
                                \cmidrule{2-4}
                                & \centering Monolingual Chinese Data & \centering ChineseWebText Dataset & \centering Cleaner Subset \arraybackslash \\
    \midrule 
    \xrowht[()]{5pt}
    \textbf{2021-43} & \centering 505.92 & \centering 187.57 & \centering 78.95 \arraybackslash\\
     \xrowht[()]{5pt}
    \textbf{2022-05} & \centering 442.47 & \centering 164.96 & \centering 69.44 \arraybackslash\\
     \xrowht[()]{5pt}
    \textbf{2022-21} & \centering 443.57 & \centering 166.75 & \centering 70.19 \arraybackslash\\
     \xrowht[()]{5pt}
    \textbf{2022-27} & \centering 417.95 & \centering 149.41 & \centering 62.70 \arraybackslash\\
     \xrowht[()]{5pt}
    \textbf{2022-33} & \centering 369.56 & \centering 123.70 & \centering 51.98 \arraybackslash\\
     \xrowht[()]{5pt}
    \textbf{2022-49} & \centering 445.29 & \centering 160.87 & \centering 67.76 \arraybackslash\\
     \xrowht[()]{5pt}
    \textbf{2023-06} & \centering 396.40 & \centering 173.47 & \centering 74.19 \arraybackslash\\
     \xrowht[()]{5pt}
    \textbf{2023-14} & \centering 441.46 & \centering 150.04 & \centering 63.33 \arraybackslash\\
     \xrowht[()]{5pt}
    \textbf{2023-23} & \centering 371.96 & \centering 143.93 & \centering 61.28 \arraybackslash\\
    \midrule\xrowht[()]{5pt}
    \textbf{Total} & \centering 3834.58 & \centering 1420.70 & \centering 599.82 \arraybackslash\\
    \bottomrule
    \end{tabular}
\end{table}
 
 \begin{table*}[htb]
\centering
     \caption{ The comparison of different pre-training datasets.}
    \label{Dataset Compare}
    \begin{tabular}{ccccc}
    \toprule
    \textbf{Dataset} & \textbf{Lang.} &\textbf{Availability}& \textbf{Pubilc Size} & \textbf{Scoring} \\
    \midrule
    C4\cite{2020T5C4}   & EN & Public &  807GB  & NO   \\  
    The Pile\cite{2020_pile} & EN & Public & 825GB  &   NO \\ 
    REFINEDWEB\cite{2023refinedweb} & EN & Public& 2.8TB    &  NO  \\
    WuDaoCorpora\cite{2021WuDaoCorpora}  & ZH & Pratly Public & 200GB  &  NO \\
    ROOTS-zh\cite{2023roots} & ZH & Public &  265GB   &  NO\\
    WanJuan1.0-zh\cite{2023wanjuan}   & ZH & Public & 550GB  & NO\\
    \textbf{ChineseWebText} (Ours)        & ZH & Public & 1.4 TB &  YES \\
    \textbf{Cleaner Subset} (Ours)        & ZH & Public & 600 GB &  YES \\
    \bottomrule
    \end{tabular}
   
\end{table*}

\subsection{Dataset Statistics and Comparison}\label{dataset_statistics_comparison}

After processing the collected CommonCrawl data with preprocessing and quality evaluation modules, this paper constructs a clean Chinese dataset ChineseWebText, which consists of 1.42 TB data. As shown in Table \ref{Data Example Format}, each text in this dataset is assigned a quality score which is generated by the quality evaluation model BERTEval. In this table, a larger quality score signifies a higher text quality. With these quality scores, LLM researchers could further select data according to a desired quality threshold. In addition to ChineseWebText, this paper also release a much cleaner subset of nearly 600 GB Chinese texts, which is built by choosing data from ChineseWebText with quality scores in the top 40\%. Through manual evaluations with three evaluators, the accuracy of this cleaner subset reaches 90\%. Table \ref{total_remain_size} shows the details of our datasets, which are extracted from nine CommonCrawl snapshots.


In Table \ref{Dataset Compare}, we compare our datasets with some other public pre-training corpora. In these work, the researchers first collect raw data from different sources, such as BookCorpus, Github, Arxiv, PubMed Central, CommonCrawl and so on. And then they clean them with some well-designed rules and algorithms. Specifically, C4\cite{2020T5C4}, The Pile\cite{2020_pile} and REFINEDWEB\cite{2023refinedweb} are three public English datasets, while WuDaoCorpora\cite{2021WuDaoCorpora}, ROOTS-zh\cite{2023roots} and WanJuan1.0-zh\cite{2023wanjuan} are three corpora for Chinese. From this table, we can see that our datasets are the latest and largest Chinese datasets. Besides, different with these previous datasets, each text in our datasets is also assigned a quality score, which could allow LLM researchers to choose data according to a new quality threshold.

\section{Data Analysis}

\subsection{Removal Rate for Different Stages}

\begin{figure}[htbp]
  \centering
  \includegraphics[width=0.95\textwidth]{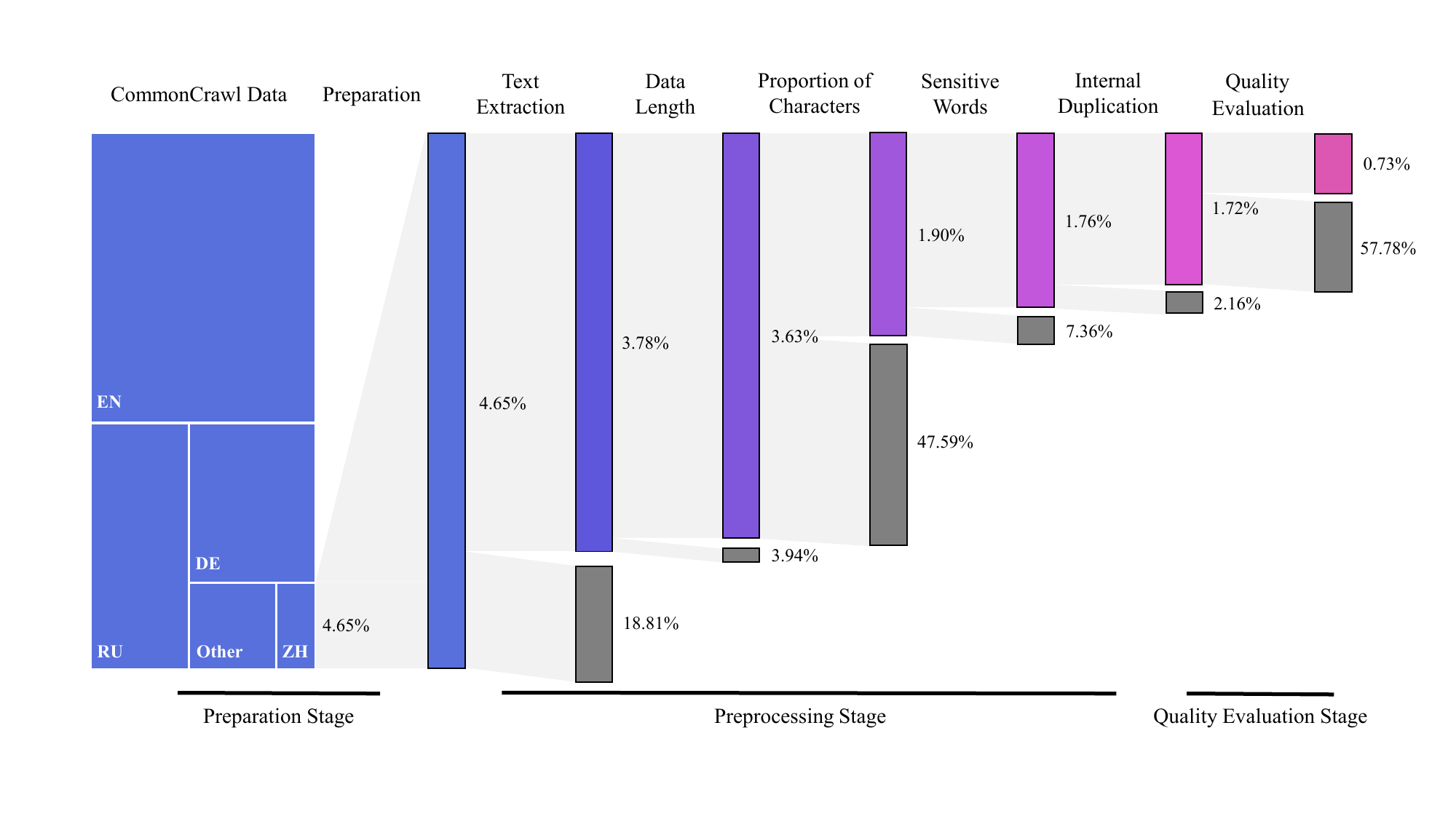}
  \caption{Removal rate for different stages. Grey represents the removal rate with respect to each previous step, while other colors represent the kept rate of all data.}
  \label{fig4removal-rate}
\end{figure}

To more precisely introduce our data processing workflow, we show Table \ref{remain_size}, which details the remaining data size and its corresponding filtering ratio for each preprocessing step and quality evaluation module. In addition, we further depict the processing workflow and the removal rate of each step in Figure \ref{fig4removal-rate}, thereby providing a high-level overview of the entire process. In each step, we show the removal ratio of data from the previous step and the absolute percentage of the remaining data from the original CommonCrawl. This facilitates readers in conveniently tracking the various processing stages from the raw data to the final data. 

Specifically, since the proportion of Chinese data is relatively low in the original CommonCrawl dataset, a large amount of data is filtered out during the preparation stage, retaining only about 4.65\% of the original data. In preprocessing stage, data is filtered in several steps. In the step of text extraction, we aim to extract all the text content and remove  redundant content generated in the preparation stage, such as useless key-value pairs. Based on this, a variety of manually defined criteria are employed to further refine the dataset, targeting the elimination of texts that either possess limited informative value, contain sensitive or inappropriate content, exhibit a low percentage of Chinese characters, or display redundant characteristics. Due to the presence of numerous entries containing traditional Chinese characters, the step of filtering based on character proportion results in a large proportion of data being cleaned up. After the preprocessing stage, we score each text of the remaining data using our evaluation model, and then construct the ChineseWebText dataset of 1.4 TB. Finally, We select the top 40\% of data based on the quality scores to construct a higher-quality subset of 600GB, which accounts for only 0.73\% of the original CommonCrawl data.

\begin{table}[htbp]
    \small
    \caption{The remaining data size and filtering ratio for each preprocessing step and quality evaluation module.}
    \label{remain_size}
    \centering
    \begin{tabular}{cp{2cm}p{1.5cm}p{1.2cm}p{2cm}p{1.5cm}p{1.5cm}p{1.5cm}}
    \toprule
    \multirow{3.2}{*}{Snapshot} & \multicolumn{7}{c}{Size After filtering operation(GB)} \\
                                \cmidrule{2-8}
                                & \centering Monolingual Chinese Data & \centering Text Extraction & \centering Data Length & \centering Proportion of Characters & \centering Sensitive Words & \centering Internal Duplication  & \centering Quality Evaluation \arraybackslash \\
    \midrule
    \textbf{2021-43} & \centering 505.92 & \centering 424.43 & \centering 409.68 & \centering 217.52 & \centering 192.84 & \centering 187.57 & \centering 78.95 \arraybackslash \\
    \xrowht[()]{7pt}
    \textit{removal rate} & \centering - & \centering \textcolor{gray}{\textit{-16.11\%}} & \centering \textcolor{gray}{\textit{-3.48\%}} & \centering \textcolor{gray}{\textit{-46.90\%}} & \centering \textcolor{gray}{\textit{-11.35\%}} & \centering \textcolor{gray}{\textit{-2.73\%}} & \centering \textcolor{gray}{\textit{-57.91\%}} \arraybackslash \\
    \textbf{2022-05} & \centering 442.47 & \centering 375.64 & \centering 362.34 & \centering 182.88 & \centering 169.01 & \centering 164.96 & \centering 69.44 \arraybackslash \\
    \xrowht[()]{7pt}
    \textit{removal rate} & \centering - & \centering \textcolor{gray}{\textit{-15.10\%}} & \centering \textcolor{gray}{\textit{-3.54\%}} & \centering \textcolor{gray}{\textit{-49.53\%}} & \centering \textcolor{gray}{\textit{-7.58\%}} & \centering \textcolor{gray}{\textit{-2.40\%}} & \centering \textcolor{gray}{\textit{-57.90\%}} \arraybackslash \\
    \textbf{2022-21} & \centering 443.57 & \centering 363.33 & \centering 348.51 & \centering 178.16 & \centering 170.09 & \centering 166.75 & \centering 70.19 \arraybackslash \\
    \xrowht[()]{7pt}
    \textit{removal rate} & \centering - & \centering \textcolor{gray}{\textit{-18.09\%}} & \centering \textcolor{gray}{\textit{-4.08\%}} & \centering \textcolor{gray}{\textit{-48.88\%}} & \centering \textcolor{gray}{\textit{-4.53\%}} & \centering \textcolor{gray}{\textit{-1.96\%}} & \centering \textcolor{gray}{\textit{-57.91\%}} \arraybackslash \\
    \textbf{2022-27} & \centering 417.95 & \centering 340.65 & \centering 326.52 & \centering 158.83 & \centering 152.33 & \centering 149.41 & \centering 62.7 \arraybackslash \\
    \xrowht[()]{7pt}
    \textit{removal rate} & \centering - & \centering \textcolor{gray}{\textit{-18.50\%}} & \centering \textcolor{gray}{\textit{-4.15\%}} & \centering \textcolor{gray}{\textit{-51.36\%}} & \centering \textcolor{gray}{\textit{-4.09\%}} & \centering \textcolor{gray}{\textit{-1.92\%}} & \centering \textcolor{gray}{\textit{-58.03\%}} \arraybackslash \\
    \textbf{2022-33} & \centering 369.56 & \centering 293.07 & \centering 280.58 & \centering 131.39 & \centering 125.84 & \centering 123.70 & \centering 51.98 \arraybackslash \\
    \xrowht[()]{7pt}
    \textit{removal rate} & \centering - & \centering \textcolor{gray}{\textit{-20.70\%}} & \centering \textcolor{gray}{\textit{-4.26\%}} & \centering \textcolor{gray}{\textit{-53.17\%}} & \centering \textcolor{gray}{\textit{-4.22\%}} & \centering \textcolor{gray}{\textit{-1.70\%}} & \centering \textcolor{gray}{\textit{-57.98\%}} \arraybackslash \\
    \textbf{2022-49} & \centering 445.29 & \centering 367.73 & \centering 352.59 & \centering 173.86 & \centering 164.34 & \centering 160.87 & \centering 67.76 \arraybackslash \\
    \xrowht[()]{7pt}
    \textit{removal rate} & \centering - & \centering \textcolor{gray}{\textit{-17.42\%}} & \centering \textcolor{gray}{\textit{-4.12\%}} & \centering \textcolor{gray}{\textit{-50.69\%}} & \centering \textcolor{gray}{\textit{-5.48\%}} & \centering \textcolor{gray}{\textit{-2.11\%}} & \centering \textcolor{gray}{\textit{-57.88\%}} \arraybackslash \\
    \textbf{2023-06} & \centering 396.40 & \centering 275.04 & \centering 263.59 & \centering 211.10 & \centering 177.44 & \centering 173.47 & \centering 74.19 \arraybackslash \\
    \xrowht[()]{7pt}
    \textit{removal rate} & \centering - & \centering \textcolor{gray}{\textit{-30.62\%}} & \centering \textcolor{gray}{\textit{-4.16\%}} & \centering \textcolor{gray}{\textit{-19.91\%}} & \centering \textcolor{gray}{\textit{-15.95\%}} & \centering \textcolor{gray}{\textit{-2.24\%}} & \centering \textcolor{gray}{\textit{-57.23\%}} \arraybackslash \\
    \textbf{2023-14} & \centering 441.46 & \centering 368.40 & \centering 354.18 & \centering 161.54 & \centering 153.27 & \centering 150.04 & \centering 63.33 \arraybackslash \\
    \xrowht[()]{7pt}
    \textit{removal rate} & \centering - & \centering \textcolor{gray}{\textit{-16.55\%}} & \centering \textcolor{gray}{\textit{-3.86\%}} & \centering \textcolor{gray}{\textit{-54.39\%}} & \centering \textcolor{gray}{\textit{-5.12\%}} & \centering \textcolor{gray}{\textit{-2.11\%}} & \centering \textcolor{gray}{\textit{-57.79\%}} \arraybackslash \\
    \textbf{2023-23} & \centering 371.96 & \centering 305.10 & \centering 292.58 & \centering 152.20 & \centering 146.90 & \centering 143.93 & \centering 61.28 \arraybackslash \\
    \xrowht[()]{7pt}
    \textit{removal rate} & \centering - & \centering \textcolor{gray}{\textit{-17.98\%}} & \centering \textcolor{gray}{\textit{-4.10\%}} & \centering \textcolor{gray}{\textit{-47.98\%}} & \centering \textcolor{gray}{\textit{-3.48\%}} & \centering \textcolor{gray}{\textit{-2.02\%}} & \centering \textcolor{gray}{\textit{-57.42\%}} \arraybackslash \\
    \midrule
    \textbf{Total} & \centering 3834.58 & \centering 3113.39 & \centering 2990.57 & \centering 1567.48 & \centering 1452.06 & \centering 1420.70 & \centering 599.82 \arraybackslash \\
    \textit{removal rate} & \centering - & \centering \textcolor{gray}{\textit{-18.81\%}} & \centering \textcolor{gray}{\textit{-3.94\%}} & \centering \textcolor{gray}{\textit{-47.59\%}} & \centering \textcolor{gray}{\textit{-7.36\%}} & \centering \textcolor{gray}{\textit{-2.16\%}} & \centering \textcolor{gray}{\textit{-57.78\%}} \arraybackslash \\
    \bottomrule
    \end{tabular}
\end{table}

\subsection{Data Quality Distribution}

To investigate the relationship between data quality and data quantity, in this section, we adopt different quality thresholds to select data from our ChineseWebText dataset. As shown in Table \ref{data quality distribution}, it presents the high-quality data size for different threshold values. In this table, the value of threshold represents the proportion of selected data in the overall dataset. Then, we hire three human evaluators to assess the quality of the selected data for each threshold. The evaluation criteria has been outlined in section 3.4.

\begin{table*}[htb]
\centering
     \caption{The data quality distribution with different quality threshold.}
    \label{data quality distribution}
    \begin{tabular}{cccccc}
    \toprule
    \multirow{2.5}{*}{\textbf{Threshold}} & \multirow{2.5}{*}{\textbf{High Quality Data Size}}&         \multicolumn{4}{c}{\textbf{Accuracy}}\\
    \cmidrule{3-6}
    & & \#1 &  \#2 & \#3 &{Average} \\
    \midrule 
    25\% & 376.90 GB & 92.80\% & 94.80\% & 94.90\% & 94.17\%  \\  
    35\% & 525.59 GB & 93.60\% & 93.60\% & 93.10\% & 93.43\% \\ 
    40\% & 599.82 GB & 90.30\% & 91.90\% & 89.50\% & 90.57\%  \\
    45\% & 672.68 GB & 84.60\% & 85.50\% & 85.59\% & 85.33\% \\
    \bottomrule
    \end{tabular}
   
\end{table*}

From Table \ref{data quality distribution},
we can observe that a lower threshold leads to higher data quality, but results in smaller data size. For example, when we keep top 25\% of our ChineseWebText, the quality can be higher than 94\%, but the remaining data only accounts for 376.9 GB. The threshold of 40\% seems to be a good choice, and it can balance the data scale and data quality. The data size can reaches about 600 GB and the average data quality with human evaluation can be higher than 90\%. Therefore, this threshold is selected to construct the cleaner subset which is released along with our ChineseWebText. Anyway, our ChineseWebText could facilitate the LLMs researchers to choose their own high-quality dataset with their desired threshold.

\subsection{Data Length Distribution}

\begin{figure}[htbp]
  \centering
  \includegraphics[width=0.6\textwidth]{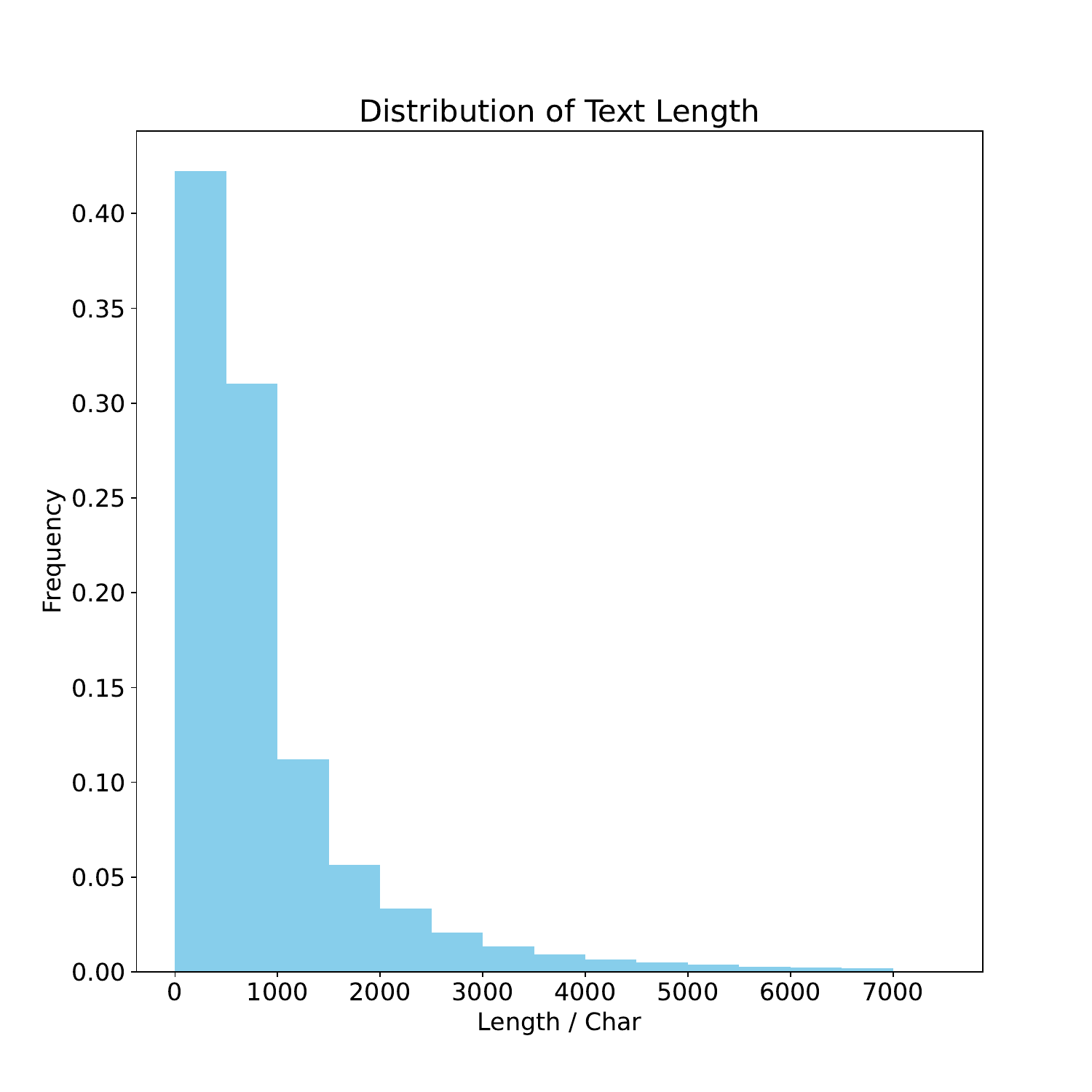}
  \caption{Length distribution of Data.}
  \label{fig4length}
\end{figure}

During the training procedure of LLMs, longer texts can provide more abundant knowledge and information, making it easier for the models to understand complex relationships in the text, and learn more knowledge. In this section, we will analyze the length distribution of the texts in our cleaner dataset. As shown in Figure  \ref{fig4length}, it illustrates the distribution of text lengths within our cleaner Chinese subset.  From this figure, we can observe that the majority of text lengths are mainly distributed within 1000 characters or less. Among them, the most significant proportion is observed within the length interval of 300 to 500 characters. Texts exceeding 1000 characters account for a relatively small portion, and there is a long tail of very long texts. After analysis, we found that the maximum text length in this dataset can reach 300,000 characters. However, they are considered to be outliers and excluded from this figure. The text length distribution in this figure could provide valuable insights into the structure and characteristics of our cleaner subset, thereby help researchers in understanding the composition of the processed dataset and facilitating the utilization of the dataset.


\section{Conclusions and Future Work}
In order to extract large-scale and high-quality Chinese pre-training data from the web, we have proposed a new pipeline approach which filter the raw crawled web data with both handcrafted rules and well-designed quality evaluation model. The rules are employed to first extract the Chinese texts and remove duplicate documents, and then filter out the explicit noisy contents such as toxic texts and advertisements. The quality evaluation model is designed based on BERT and can assign each text with a quality score. With the proposed approach, we release the latest and largest Chinese dataset of 1.4 TB, each of which is associated with a quality score, facilitating the LLMs researchers to re-filter the data with desired quality thresholds. We further release a much cleaner subset of 600 GB Chinese data with the quality exceeding 90\% by human evaluation. We also release the complete tool-chain that processes the raw data into the clean texts.

In the future, we will continue to enlarge the Chinese dataset with newly incoming web data. Meanwhile, we are going to explore better algorithms and strategies for data filtering. For example, we can design quality evaluation models for each kind of data noise.

\bibliographystyle{unsrt}

\bibliography{references}  






\end{document}